\documentclass{article}
\usepackage{fullpage}
\usepackage{times}
\usepackage{helvet}
\usepackage{courier}
\usepackage{amssymb,amsmath}
\usepackage{comment}
\usepackage{graphicx}

\usepackage{float}
\usepackage[boxed,noend]{algorithm2e}
\SetKwInput{Input}{Input}
\SetKw{KwReject}{reject}
\SetKw{KwAccept}{accept}
\SetKw{KwReturn}{return}

\newcommand{\ignore}[1]{}

\begin{document}

\title{Arc Consistency and Friends}
\author{Hubie Chen\footnote{
Chen is supported by the Spanish program ``Ramon y Cajal''.
}, Victor Dalmau\footnote{
Both Chen and Dalmau are supported by 
MICINN grant TIN2010-20967-C04-02.
}, Berit Gru\ss ien\footnote{
Work by Gru\ss ien was supported by the Deutsche Forschungsgemeinschaft (DFG)
within the research training group "Methods for Discrete Structures"
(GrK 1408).
}}


\maketitle

\begin{abstract}
\begin{quote}
A natural and established way to restrict the constraint satisfaction problem
is to fix the relations that can be used to pose constraints; 
such a family of relations is called a \emph{constraint language}.
In this article, we study arc consistency, a heavily investigated inference method,
and three extensions thereof from the perspective of constraint languages.
We conduct a comparison of the studied methods on the basis of 
which constraint languages they solve, and we present new polynomial-time
tractability
results for singleton arc consistency, the most powerful method studied.
\end{quote}
\end{abstract}

\newtheorem{theorem}                            {Theorem}
\newtheorem{lemma}              [theorem]       {Lemma}
\newtheorem{remark}             [theorem]       {Remark}
\newtheorem{fact}               [theorem]       {Fact}
\newtheorem{claim}              [theorem]       {Claim}
\newtheorem{corollary}          [theorem]       {Corollary}
\newtheorem{definition}         [theorem]       {Definition}
\newtheorem{prop}               [theorem]       {Proposition}
\newtheorem{warning}            [theorem]       {Warning}
\newtheorem{observation}  						    {Observation}
\newtheorem{example}        [theorem]       {Example}

\newenvironment{pf}[1][]{\noindent\textbf{Proof\/}#1\/. }{$\Box$ \vspace{1mm}}

\newcommand{\rats}{\mathbb{Q}}

\newcommand{\rela}{\mathbf{A}}
\newcommand{\relb}{\mathbf{B}}
\newcommand{\relc}{\mathbf{C}}
\newcommand{\reld}{\mathbf{D}}
\newcommand{\relg}{\mathbf{G}}
\newcommand{\pow}{\wp}

\newcommand{\csp}{{\mathsf{CSP}}}

\newcommand{\fancya}{\mathcal{A}}

\newcommand{\ind}{\mathrm{Ind}}

\newcommand{\homo}{\rightarrow}

\newcommand{\tup}[1]{\overline{#1}}

\newcommand{\alga}{\mathbb{A}}
\newcommand{\algb}{\mathbb{B}}
\newcommand{\algab}{\mathbb{A}_{\relb}}
\newcommand{\algg}{\mathbb{G}}

\newcommand{\pol}{\mathsf{Pol}}

\section{Introduction}

\subsection{Background}

The constraint satisfaction problem (CSP) involves deciding,
given a set of variables and a set of constraints on the variables,
whether or not there is an assignment to the variables satisfying all of the constraints.
Cases of the constraint satisfaction problem appear in many fields of study,
including artificial intelligence, spatial and temporal reasoning,
logic, combinatorics, and algebra.
Indeed, the constraint satisfaction problem is flexible in that 
it admits a number of equivalent formulations. 
In this paper, we work with the formulation as the
relational homomorphism problem: given two similar relational structures
$\rela$ and $\relb$, does there exist a homomorphism from $\rela$ to $\relb$?
In this formulation, 
one can view
each relation of $\rela$ as containing  variable tuples
that are constrained together, and the corresponding relation of $\relb$
as containing the permissible values for the variable tuples~\cite{FederVardi}.

The constraint satisfaction problem is in general NP-hard;
this general intractability has motivated the study of restricted versions of
the CSP that have various desirable complexity and algorithmic properties.
A natural and well-studied way to restrict the CSP is to fix the value relations
that can be used to pose constraints;
in the homomorphism formulation,
this corresponds to fixing the right-hand side structure $\relb$,
which is also known as the \emph{constraint language}.
Each structure $\relb$ then gives rise to a problem $\csp(\relb)$,
and one obtains a rich family of problems that include
boolean satisfiability problems, graph homomorphism problems, and
satisfiability problems on algebraic equations.
One of the primary current research threads 
involving such problems
is to understand for which
finite-universe constraint languages $\relb$
the problem $\csp(\relb)$ is polynomial-time
tractable~\cite{bulatov-valeriote}; 
there is also work on characterizing the languages
$\relb$ for which the problem $\csp(\relb)$ is contained in lower complexity
classes such as L (logarithmic space) and
NL (non-deterministic logarithmic space)~\cite{dalmau-pathwidth,larose-tesson}.  
With such aims providing motivation, there have been
efforts to characterize the languages amenable
to solution by certain algorithmic techniques, notably,
representing solution spaces by generating sets~\cite{IMMVW}
and consistency methods~\cite{LaroseZadori,abd-affine,bkbw}, which we now turn to discuss.

Checking for \emph{consistency} is
a primary reasoning technique for the practical solution of the CSP,
and has been studied theoretically from many 
viewpoints~\cite{LaroseZadori,abd-affine,akv,abd,aft,bkbw,atserias-weyer}.
The most basic and simplest form of consistency is \emph{arc consistency},
which algorithmically involves performing inferences concerning the set of feasible
values for each variable.  The question of how to efficiently implement 
an arc consistency check has been studied intensely, and highly optimized
implementations that are linear in both time and space have been
presented.
In general, a consistency check typically involves running an efficient method
that performs inference on bounded-size sets of variables, and which
can sometimes detect that a CSP instance is 
inconsistent and has no solution.
While these methods exhibit one-sided error in that they
do not catch all non-soluble CSP instances (as one expects from
the conjunction of their efficiency and the intractability of the CSP),
it has been shown that, for certain constraint languages,
they can serve as complete decision procedures, by which is meant,
they detect an inconsistency if (and only if) an instance has no solution.
As an example, \emph{unit propagation}, a consistency method
that can be viewed as arc consistency specialized to SAT formulas,
is well-known to decide the Horn-SAT
problem in this sense.

\subsection{Contributions}

In this paper, we study arc consistency and three natural extensions 
thereof from the perspective of constraint languages.
The extensions of AC that we study are 
look-ahead arc consistency (LAAC)~\cite{SLAAC};
peek arc consistency (PAC)~\cite{peek},
and singleton arc consistency (SAC)~\cite{db-sac,bd-theoretical}.
Each of these algorithms is natural, conceptually simple, readily understandable,
and easily implementable using arc consistency as a black box.
Tractability results for constraint languages have been presented
for AC by
Feder and Vardi~\cite{FederVardi} (for instance); 
and for LAAC and PAC in the previously cited work.
In fact,
for each of these 
three algorithms, characterizations of the class of tractable languages
have been given, as we discuss in the paper.

We give a uniform presentation of these algorithms (Section~\ref{sect:algorithms}), and 
conduct a comparison of these algorithms 
on the basis of which languages they solve (Section~\ref{sect:comparison}).
Our comparison shows, roughly, that the algorithms
can be placed into a hierarchy:
solvability of a language by AC or LAAC implies solvability by PAC;
solvability by PAC in turn implies solvability by SAC
(see Section~\ref{sect:comparison} for precise statements).
We also study the strictness of the containments shown.
We thus contribute to a basic, foundational understanding 
of the scope of these algorithms and of the situations in which these algorithms
can be demonstrated to be effective.

We then present new tractability results for singleton arc consistency 
(Section~\ref{sect:tractability}).
We prove that languages having certain types of 
\emph{2-semilattice polymorphisms} can be solved by singleton arc consistency;
and, we prove that any language having a \emph{majority polymorphism}
is solvable by singleton arc consistency.
The presence of a majority polymorphism is a robust and well-studied condition:
majority polymorphisms were used to give some of the initial 
language tractability results, are known to exactly characterize the languages
such that \emph{3-consistency} implies \emph{global consistency}
(we refer to~\cite{CCC} for definitions and more details), 
and gave one of the first large classes
of languages whose constraint satisfaction problem 
could be placed in non-deterministic
logarithmic space~\cite{majority-nl}.
While the languages that we study are already known to be polynomial-time
tractable~\cite{CCC,twosemilattices},
from the standpoint of understanding the complexity
and algorithmic properties of constraint languages,
we believe our tractability results to be particularly attractive for a couple of reasons.
First, relative to a fixed language, singleton arc consistency runs in 
quadratic time~\cite{bd-theoretical},
constituting a highly non-trivial running time improvement
over the cubic time bound that was previously known for
the studied languages.
Also, in showing that these languages are amenable to solution by
singleton arc consistency, we demonstrate their polynomial-time tractability 
in an alternative fashion via an algorithm that is different from the
previously used ones; 
the techniques that we employ expose a different type of structure
in the studied constraint languages.




\section{Preliminaries}
Our definitions and notation are fairly standard.
For an integer $k \geq 1$, we use the notation $[k]$ to
denote the set containing the first $k$ positive integers,
that is, the set $\{ 1, \ldots, k \}$.

\newcommand{\pr}{\pi}

\paragraph{Structures.}
A \emph{tuple} over a set $B$ is an element of $B^k$
for a value $k \geq 1$ called the \emph{arity} of the tuple;
when $\tup{t}$ is a tuple, we often use the notation $\tup{t} = (t_1, \ldots, t_k)$
to denote its entries.
A \emph{relation} over a set $B$ is a subset of $B^k$ 
for a value $k \geq 1$ called the \emph{arity} of the relation.
We use $\pi_i$ to denote the operator that projects onto the $i$th coordinate:
$\pi_i(\tup{t})$ denotes the $i$th entry $t_i$ of a tuple
$\tup{t} = (t_1, \ldots, t_k)$, and for a relation $R$ we define
$\pi_i(R) = \{ \pi_i(\tup{t}) ~|~ \tup{t} \in R \}$.
Similarly, for a subset $I \subseteq [k]$
whose elements are 
$i_1 < \cdots < i_m$, 
we use
$\pi_I(\tup{t})$ to denote the tuple
$(t_{i_1}, \ldots, t_{i_m})$, and we define
$\pi_I(R) = \{ \pi_I(\tup{t}) ~|~ \tup{t} \in R \}$.

A \emph{signature} $\sigma$ is a set of symbols, each of which
has an associated arity.  
A \emph{structure} $\relb$ over signature $\sigma$
consists of a universe $B$, which is a set,
and a relation $R^{\relb} \subseteq B^k$ for each symbol $R \in \sigma$ of arity $k$.
(Note that in this paper, we are concerned only with relational structures,
which we refer to simply as structures.)
Throughout, we will use the bold capital letters $\rela, \relb, \ldots$
to denote structures, and the corresponding non-bold capital letters
$A, B, \ldots$ to denote their universes.
We say that a structure $\relb$ is \emph{finite} if its universe
$B$ has finite size.
Unless stated otherwise, we assume all structures under discussion 
in this paper to be finite.
We say that a structure $\relb$ \emph{has all constants}
if for each $b \in B$, there is a relation symbol $R_b$ with
$R_b^{\relb} = \{ (b) \}$.

When two structures $\rela$ and $\relb$ are defined over the same signature
$\sigma$, we say that they are \emph{similar}.  
We define the following notions on similar structures.
For similar structures $\rela$ and $\relb$ over a signature $\sigma$,
we say that $\rela$
is an \emph{induced substructure}
of $\relb$ if $A \subseteq B$
and for every $R \in \sigma$ of arity $k$, it holds that
$R^{\rela} = A^k \cap R^{\relb}$.
Observe that for a structure $\relb$ and a subset $B' \subseteq B$,
there is exactly one induced substructure of $\relb$ with universe $B'$.
For similar structures $\rela$ and $\relb$ over a signature $\sigma$,
the product structure $\rela \times \relb$ is defined to be
the structure with universe $A \times B$ and such that
$R^{\rela \times \relb} = 
\{ ((a_1, b_1), \ldots, (a_k, b_k)) ~|~ \tup{a} \in R^{\rela}, \tup{b} \in R^{\relb} \}$ for all $R \in \sigma$.
We use $\rela^n$ to denote the $n$-fold product 
$\rela \times \cdots \times \rela$.

We say that a structure $\relb$ 
over signature $\sigma'$
is an \emph{expansion} of another
structure $\rela$ over signature $\sigma$
if (1) $\sigma' \supseteq \sigma$,
(2) the universe of $\relb$ is equal to the universe of $\rela$,
and 
(3) for every symbol $R \in \sigma$, it holds that
$R^{\relb} = R^{\rela}$.
We will use the following non-standard notation.
For any structure $\rela$ (over signature $\sigma$)
and any subset $S \subseteq A$,
we define $[\rela, S]$ to be the expansion of $\rela$
with the signature
$\sigma \cup \{ U \}$
where $U$ is a new symbol of arity $1$, defined by
$U^{[\rela, S]} = S$ and 
$R^{[\rela, S]} = R^{\rela}$ for all $R \in \sigma$.
More generally, for a structure $\rela$ (over $\sigma$)
and a sequence of subsets $S_1, \ldots, S_n \subseteq A$,
we define $[\rela, S_1, \ldots, S_n]$ to be the expansion of $\rela$
with the signature
$\sigma \cup \{ U_1, \ldots, U_n \}$
where $U_1, \ldots, U_n$ are new symbols of arity $1$, defined by
$U_i^{[\rela, S_1, \ldots, S_n]} = S_i$ for all $i \in [n]$,
and
$R^{[\rela, S_1, \ldots, S_n]} = R^{\rela}$ for all $R \in \sigma$.

\paragraph{Homomorphisms and the constraint satisfaction problem.}
For similar structures $\rela$ and $\relb$ over the signature $\sigma$,
a \emph{homomorphism} from $\rela$ to $\relb$
is a mapping $h: A \rightarrow B$ such that
for every symbol $R$ of $\sigma$ and every tuple
$(a_1, \ldots, a_k) \in R^{\rela}$, it holds that
$(h(a_1), \ldots, h(a_k)) \in R^{\relb}$.
We use $\rela \homo \relb$ to indicate that there is a homomorphism
from $\rela$ to $\relb$; 
when this holds, we also say that $\rela$ \emph{is homomorphic to} $\relb$.
It is well-known and straightforward to verify that
the homomorphism relation $\homo$ is transitive, that is,
if $\rela \homo \relb$ and $\relb \homo \relc$, then $\rela \homo \relc$.

The \emph{constraint satisfaction problem (CSP)} is the problem of deciding,
given as input a pair $(\rela, \relb)$ of similar structures,
whether or not there exists a homomorphism from $\rela$
to $\relb$.  
When $(\rela, \relb)$ is an instance of the CSP,
we will also call a homomorphism from $\rela$ to $\relb$
a \emph{satisfying assignment}; say that the instance is \emph{satisfiable}
if there exists such a homomorphism; and,
say that the instance is \emph{unsatisfiable} if there does not exist such a homomorphism.
We generally assume that in an instance of the CSP,
the left-hand side structure $\rela$ contains finitely many tuples.
For any structure $\relb$ (over $\sigma$), 
the \emph{constraint satisfaction problem for $\relb$}, denoted by $\csp(\relb)$, 
is the constraint satisfaction problem where the right-hand side structure
is fixed to be $\relb$, that is, the problem of
deciding, given as input a structure $\rela$ over $\sigma$,
whether or not
there exists a homomorphism from $\rela$ to $\relb$.
In discussing a problem of the form 
$\csp(\relb)$, the structure $\relb$ is often referred to as
the \emph{template} or \emph{constraint language}.
There are several equivalent definitions of the constraint satisfaction
problem.
For instance,
in logic, the constraint satisfaction problem can be formulated as 
the model checking problem for primitive positive sentences over relational structures,
and in database theory, it can be formulated 
as the containment problem for conjunctive queries~\cite{ChandraMerlin}. 

\paragraph{Polymorphisms.}
When $f: B^n \rightarrow B$ is an operation on $B$
and 
$$\tup{t_1} = (t_{11}, \ldots, t_{1k}), \ldots, \tup{t_n} = (t_{n1}, \ldots, t_{nk}) \in B^k$$
are tuples of the same arity $k$ 
over $B$, we use
$f(\tup{t_1}, \ldots, \tup{t_n})$ to denote the arity $k$ tuple
obtained by applying $f$ coordinatewise, that is, 
$$f(\tup{t_1}, \ldots, \tup{t_n}) = 
(f(t_{11}, \ldots, t_{n1}), \ldots, f(t_{1k}, \ldots, t_{nk})).$$
An operation $f: B^n \rightarrow B$ is a \emph{polymorphism} 
of a structure $\relb$ over $\sigma$ if for every symbol $R \in \sigma$
and any tuples $\tup{t_1}, \ldots, \tup{t_n} \in R^{\relb}$, 
it holds that $f(\tup{t_1}, \ldots, \tup{t_n}) \in R^{\relb}$.
That is, each relation $R^{\relb}$ is closed under the action of $f$.
Equivalently, an operation $f: B^n \rightarrow B$ 
is a polymorphism of $\relb$ if
it is a homomorphism from $\relb^n$ to $\relb$.

\section{Algorithms}
\label{sect:algorithms}

In this section, we 
give a uniform presentation of the
four algorithms under investigation in this paper:
arc consistency, look-ahead arc consistency, peek arc consistency, 
and singleton arc consistency,
presented in Sections~\ref{ss:ac}, \ref{ss:laac}, \ref{ss:pac},
and \ref{ss:sac}, respectively.
The results on the first three algorithms come from previous work,
as we discuss in presenting each of these algorithms; 
for singleton arc consistency, we here develop results similar to those
given for the other algorithms.

Our treatment of arc consistency, peek arc consistency, and singleton arc consistency
is uniform: for each of these algorithms, 
we present a homomorphism-based consistency condition,
we show that the algorithm checks precisely this consistency condition,
and we give an algebraic condition describing the structures $\relb$
such that the algorithm solves $\csp(\relb)$.
These three algorithms give one-sided consistency checks: 
each either correctly rejects an instance as unsatisfiable
or outputs ``?'', which can be interpreted as a report
that it is unknown whether or not the instance is satisfiable.
The other algorithm, look-ahead arc consistency, has a somewhat different character.
It attempts to build a satisfying assignment one variable at a time,
using arc consistency as a filtering criterion; 
it either returns a satisfying assignment, or outputs ``?''.

Throughout this section and in later sections, we will make use of 
a structure $\pow(\relb)$ that is defined for every structure $\relb$,
as follows~\cite{FederVardi,DalmauPearson}.  
For a structure $\relb$ (over $\sigma$),
we define
$\pow(\relb)$ to be the structure with universe
$\pow(B) \setminus \{ \emptyset \}$ and where,
for every symbol $R \in \sigma$ of arity $k$,
$R^{\pow(\relb)} = \{ (\pi_1 S, \ldots, \pi_k S) ~|~ S \subseteq R^{\relb}, S \neq \emptyset \}$.
Here, $\pow(B)$ denotes the power set of the set $B$.

\subsection{Arc Consistency}
\label{ss:ac}

We now present the arc consistency algorithm.  
The main idea of the algorithm is to associate to each element $a \in A$
a set $S_a$ of values which, 
throughout the execution of the algorithm,
has the property that for any solution $h$,
it must hold that $h(a) \in S_a$.
The algorithm continually shrinks the sets
$S_a$ in a natural fashion until they stabilize; at this point,
if some set $S_a$ is the empty set, then no solution can exist, and the
algorithm rejects the instance.

\begin{algorithm}[H]
{\bfseries Arc Consistency}\\
\Input{a pair $(\rela,\relb)$ of similar structures}
\BlankLine 
\hrule
\BlankLine
\ForAll{$a\in A$}{set $S_a:=B$\nllabel{AC2}\;}
\Repeat{no set $S_a$ is changed$\!$}
{
	\ForAll{relations $R^\rela$ of $\rela$}{
		\ForAll{tuples $(a_1,\ldots ,a_k)\in R^\rela$}{
			\ForAll{$i\in[k]$}{
				set $S_{a_i}:=\pi_i( R^\relb \cap (S_{a_1}\times \ldots \times S_{a_k}))$\nllabel{AC7}\;
			}
		}
	}
}
\lIf{	there exists $a\in A$ such that $S_{a}=\emptyset$}{
	\KwReject$\!$\;
}
\lElse{
	\KwReturn ``?''\;
}
\end{algorithm}

Feder and Vardi~\cite{FederVardi} have studied arc consistency, 
under an equivalent
formulation in terms of Datalog Programs, for constraint languages.
The results in this section are due to this reference. The connection
of the results in Feder and Vardi with arc consistency was made explicit in Dalmau
and Pearson~\cite{DalmauPearson}.

\begin{definition}
\label{def:acc}
An instance $(\rela, \relb)$ has the
\emph{arc consistency condition (ACC)} if
there exists a homomorphism from $\rela$ to $\pow(\relb)$.
\end{definition}

\begin{prop}
\label{prop:acc}
The arc consistency algorithm does not reject 
an instance $(\rela, \relb)$
if and only if
the instance has the ACC.
\end{prop}


\begin{definition}
\label{def:ac-solves}
Let $\relb$ be a structure.
We say that arc consistency \emph{solves} $\csp(\relb)$
if for all structures $\rela$, the following holds:
$(\rela, \relb)$ has the ACC implies that 
there is a homomorphism $\rela \rightarrow \relb$.
\end{definition}

Note that the converse of the condition given in this definition
always holds: if $h$ is a homomorphism from $\rela$ to $\relb$,
then the mapping sending each $a \in A$ to the set $\{ h(a) \}$
is a homomorphism from $\rela$ to $\pow(\relb)$.

\begin{theorem} 
\label{thm:ac-solves}
Let $\relb$ be a structure.
Arc consistency solves $\csp(\relb)$ if and only if
there is a homomorphism $\pow(\relb) \rightarrow \relb$.
\end{theorem}

\subsection{Look-Ahead Arc Consistency}
\label{ss:laac}

We now present the look-ahead arc consistency algorithm.
It attempts to construct a satisfying assignment by setting one variable at a time,
using arc consistency as a filter to find a suitable value for each variable.

\begin{algorithm}[H]
{
\bfseries Look-Ahead Arc Consistency}\\
\Input{a pair $(\rela,\relb)$ of similar structures}
\BlankLine 
\hrule
\BlankLine
\ForAll{$a\in A$}{set $S_a:=B$\;}
\For{$i=1$ \KwTo $|A|$}{
	pick arbitrary $a_i\in A$ with $a_i\not \in \{a_1,\ldots ,a_{i-1}\}$\;
	\ForAll{$b\in B$}{
		\If{ {Arc Consistency} $([\rela, \{a_1 \},\ldots , \{ a_{i-1} \},  \{ a_i \}],
			[\relb,\{b_1 \},\ldots, \{b_{i-1} \}, \{ b \}])$ rejects}{
			remove $b$ from $S_{a_i}$\;	
		}
	}
	\lIf{$S_{a_i}=\emptyset$}{\KwReturn ``?''\;}
	\lElse{choose $b_i\in S_{a_i}$ arbitrarily\;}
}
\KwAccept\;
\end{algorithm}

Look-ahead arc consistency was introduced and studied by
Chen and Dalmau~\cite{SLAAC}, and the theorem that follows is due to them.
This algorithm can be viewed as a generalization
of an algorithm for SAT studied by Del Val~\cite{delval}.

\begin{definition}
\label{def:laac-solves}
Let $\relb$ be a structure.
We say that look-ahead arc consistency \emph{solves} $\csp(\relb)$
if for all structures $\rela$, the following holds:
if there exists a homomorphism $\rela \rightarrow \relb$, then
the look-ahead arc consistency algorithm, given $(\rela, \relb)$, 
outputs such a homomorphism.
\end{definition}

\begin{theorem}
\label{thm:laac-solves}
Let $\relb$ be a structure.
Look-ahead arc consistency solves $\csp(\relb)$ 
if and only if there is a homomorphism 
$l: \pow(\relb) \times \relb \rightarrow \relb$ such that
$l( \{ b \}, b') = b$ for all $b, b' \in B$.
\end{theorem}

\subsection{Peek Arc Consistency}
\label{ss:pac}

We now present the peek arc consistency algorithm.
It attempts to find, for each variable $a \in A$, a value $b \in B$
such that when $a$ is set to $b$, the arc consistency check is passed.

\begin{algorithm}[H]
{\bfseries Peek Arc Consistency}\\
\Input{a pair $(\rela,\relb)$ of similar structures}
\BlankLine 
\hrule
\BlankLine
\ForAll{$a\in A$}{set $S_a:=B$\;}
\ForAll{$a\in A,b\in B$}{
	\If{Arc Consistency $([\rela, \{a\}],[\relb,\{b\}])$ rejects}{
		remove $b$ from $S_a$\;	
	}
}
\lIf{there exists $a\in A$ such that $S_{a}=\emptyset$}{
	\KwReject$\!$\;
}
\lElse{
	\KwReturn ``?''\;	
}

\end{algorithm}

Peek arc consistency was introduced and studied by
Bodirsky and Chen~\cite{peek}; the notions and results that follow come from them.
In their work, the algorithm is shown to solve certain constraint languages,
including some languages having infinite-size universes; such languages
actually gave the motivation for introducing the algorithm.
In this work, it is pointed out that peek arc consistency can be readily parallelized;
by invoking the  arc consistency checks independently in parallel, 
one can achieve a linear parallel running time.

\begin{definition}
\label{def:pacc}
An instance $(\rela, \relb)$ has the
\emph{peek arc consistency condition (PACC)} if
for every element $a \in A$,
there exists a homomorphism $h$ from $\rela$ to $\pow(\relb)$
such that $h(a)$ is a singleton.
\end{definition}

\begin{prop}
\label{prop:pacc}
The peek arc consistency algorithm does not reject an instance
$(\rela, \relb)$ if and only if the instance has the PACC.
\end{prop}

\begin{definition}
\label{def:pac-solves}
Let $\relb$ be a structure.
We say that peek arc consistency \emph{solves} $\csp(\relb)$
if for all structures $\rela$, the following holds:
$(\rela, \relb)$ has the PACC implies that 
there is a homomorphism $\rela \rightarrow \relb$.
\end{definition}

The converse of the condition given in this definition always holds.
Suppose that 
$h$ is a homomorphism from $\rela$ to $\relb$;
then,
the mapping taking each $a \in A$ to the singleton $\{ h(a) \}$
is a homomorphism from $\rela$ to $\pow(\relb)$
and hence $(\rela, \relb)$ has the PACC.

\newcommand{\sing}{\mathsf{Sing}}
\newcommand{\unionsing}{\mathsf{UnionSing}}

We use the notation $\sing(\pow(\relb)^n)$ to denote
the induced substructure of $\pow(\relb)^n$ whose universe
contains an $n$-tuple of $\pow(\relb)^n$ if and only if
at least one coordinate of the tuple is a singleton.

\begin{theorem} \label{thm:pac-solves}
Let $\relb$ be a structure.  Peek arc consistency solves $\csp(\relb)$
if and only if for all $n \geq 1$
there is a homomorphism $\sing(\pow(\relb)^n) \rightarrow \relb$.
\end{theorem}

\subsection{Singleton Arc Consistency}
\label{ss:sac}

We now present the singleton arc consistency algorithm.
As with arc consistency, this algorithm associates to each element $a \in A$
a set $S_a$ of feasible values.  It then continually checks, for pairs $(a, b)$
with $a \in A$ and $b \in S_a$, whether or not arc consistency can be established
with respect to the sets $S_a$ and when $a$ is assigned to $b$; 
if for some pair $(a, b)$
it cannot, then $b$ is removed from the set $S_a$.
As with arc consistency, this algorithm's outer loop runs until the sets $S_a$
stabilize, and the algorithm rejects if one of the sets $S_a$ is equal to the empty set.

\begin{algorithm}[H]
{\bfseries Singleton Arc Consistency}\\
\Input{a pair $(\rela,\relb)$ of similar structures}
\BlankLine 
\hrule
\BlankLine
\ForAll{$a\in A$}{set $S_a:=B$\;}
denote $A=\{a_1,\ldots ,a_n\}$\;
\Repeat{no set $S_a$ is changed$\!$}
{	
	\ForAll{$a\in A,b\in S_a$}{
		\If{ Arc Consistency
			$([\rela , \{a_1\},\ldots,\{a_n\},\{a\}],[\relb, S_{a_1},\ldots, S_{a_n},\{b\}])$
			rejects 
			}{
			remove $b$ from $S_a$\;
		}
	}
}
\lIf{there exists $a\in A$ such that $S_{a}=\emptyset$}{
	\KwReject$\!$\;
}
\lElse{
	\KwReturn ``?''\;
}
\end{algorithm}

Singleton arc consistency was introduced by 
Debruyne and Bessiere~\cite{db-sac}.
We now give a development of singleton arc consistency
analogous to that of arc consistency and peek arc consistency.

\begin{definition}
\label{def:sacc}
An instance $(\rela, \relb)$ has the
\emph{singleton arc consistency condition (SACC)} if
there exists a mapping ${s: A \rightarrow \pow(B) \setminus \{ \emptyset \}}$
such that for all $a \in A$, $b \in s(a)$
there exists a homomorphism $h_{a,b}: \rela \rightarrow \pow(\relb)$
where: 

\begin{itemize}

\item  $h_{a,b}(a)=\{b\}$, and

\item  for all $a' \in A$, it holds that $h_{a,b}(a') \subseteq s(a')$.
\end{itemize}
\end{definition}

\begin{prop}
\label{prop:sacc}
The singleton arc consistency algorithm does not reject an instance
$(\rela, \relb)$ if and only if the instance has the SACC.
\end{prop}

\begin{pf}
Suppose that the singleton arc consistency algorithm does not reject
an instance $(\rela, \relb)$.
Let $\{ S_a \}_{a \in A}$ denote the sets computed by the algorithm
at the point of termination, and
define $s$ to be the mapping where $s(a) =S_a$ for all $a\in A$.
Let $a \in A$ and $b \in s(a)$.  By the definition of the algorithm,
the pair
$([\rela , \{a_1\},\ldots,\{a_n\},\{a\}],\allowbreak[\relb, S_{a_1},\ldots, S_{a_n},\{b\}])$
has the ACC, and thus the desired homomorphism $h_{a,b}$ exists.

Now, suppose that the instance $(\rela, \relb)$ has the SACC, and let
$s$ be a mapping with the described properties.
We show that throughout the execution of the algorithm,
it holds that $s(a)\subseteq S_a$ for all $a\in A$.
First, $S_a$ is initialized with $B$ for every $a\in A$. 
Next, we show that when $a \in A$ and $b \in s(a)$,
then $b$ is never removed from $S_a$ by the algorithm.
This is because
by definition of SACC,
there exists a homomorphism $h_{a,b}: \rela \rightarrow \pow(\relb)$ with $h_{a,b}(a)=\{b\}$
such that for all $a' \in A$, it holds that $h_{a,b}(a') \subseteq s(a')$. 
Since $s(a')\subseteq S_{a'}$ by the inductive assumption,
$([\rela , \{a_1\},\ldots,\{a_n\},\{a\}],[\relb, S_{a_1},\ldots, S_{a_n},\{b\}])$
has the ACC and hence the algorithm does not
remove $b$ from $S_a$.
\end{pf}

\begin{definition}
\label{def:sac-solves}
Let $\relb$ be a structure.
We say that singleton arc consistency \emph{solves} $\csp(\relb)$
if for all structures $\rela$, the following holds:
$(\rela, \relb)$ has the SACC implies that 
there is a homomorphism $\rela \rightarrow \relb$.
\end{definition}

The converse of the condition given in this definition always holds:
suppose that $h$ is a homomorphism from $\rela$ to $\relb$.
Then, the instance $(\rela, \relb)$ has the SACC
via the mapping $s$ where $s(a) = \{ h(a) \}$ for all $a \in A$
and the mappings $h_{a, b}$ defined by
$h_{a, b}(a') = \{ h(a') \}$ for all $a' \in A$.

We use the notation $\unionsing(\pow(\relb)^n)$ to denote
the induced substructure of $\pow(\relb)^n$ whose universe
contains an $n$-tuple $(S_1, \ldots, S_n)$
of $\pow(\relb)^n$ if and only if
it holds that $\bigcup_{i \in [n]} S_i = \bigcup_{i \in [n], |S_i| = 1} S_i$.

\begin{theorem} 
\label{thm:sac-solves}
Let $\relb$ be a structure.  Singleton arc consistency solves $\csp(\relb)$
if and only if for all $n \geq 1$ there is a homomorphism
$\unionsing(\pow(\relb)^n) \rightarrow \relb$.
\end{theorem}

\begin{pf}
First we show that if singleton arc consistency solves 
$\csp(\relb)$, then
there is a homomorphism from \linebreak $\unionsing(\pow(\relb)^{n})$ to $\relb$ for all $n\geq 1$.
Let $n \geq 1$; 
we show that $(\unionsing(\pow(\relb)^{n}),\relb)$ has the SACC.
Then, there is a homomorphism from $\unionsing(\pow(\relb)^{n})$ to $\relb$,
since the singleton arc consistency algorithm solves {\upshape CSP($\relb$)}.

Let $s$ be the mapping $s(a):=\bigcup_{i \in [n]} S_i$ for all tuples
$a=(S_1,\dots,S_n)$ of $\unionsing(\pow(\relb)^{n})$.
Now let us consider an arbitrary tuple 
$a=(S_1,\dots,S_n)$ of $\unionsing(\pow(\relb)^{n})$ and an arbitrary $b \in s(a)$.
Since $\bigcup_{i \in [n]} S_i = \bigcup_{i \in [n], |S_i| = 1} S_i$, 
there is an $i\in[n]$ such that $\{b\}=S_i$.
Thus, the homomorphism $\pi_i\colon \unionsing(\pow(\relb)^{n}) \rightarrow \pow(\relb)$ 
that projects onto the $i$th coordinate
satisfies $\pi_i(a)=\{b\}$,
and for all tuples $a'$ of $\unionsing(\pow(\relb)^{n})$, it holds that
$\pi_i(a')\subseteq s(a')$. Hence, $(\unionsing(\pow(\relb)^{n}),\relb)$ has the SACC.

For the other direction, we show that
if there is a homomorphism from $\unionsing(\pow(\relb)^{n})$ to $\relb$ for all $n \geq 1$,
then singleton arc consistency solves $\csp(\relb)$.
Thus, we have to show that there exists a homomorphism from $\rela$ to $\relb$
if $(\rela,\relb)$ has the SACC.
Let $s$ be the homomorphism from the definition of SACC, and let us use
$\{h_1,\ldots ,h_n\}$ to denote the set 
${\{h_{a,b}\mid a\in A,b\in s(a)\}}$ of homomorphisms.
Further, let $g$ be the homomorphism $(h_1,\ldots ,h_n) \colon \rela\to \pow(\relb)^{n}$.
Now, for every element $a\in A$ the image $g(a)=(h_1(a),\ldots ,h_n(a))$ is a tuple of $\unionsing(\pow(\relb)^{n})$:
for every $b\in \bigcup_{j \in [n]} h_j(a)$,
it holds that $b \in s(a)$ and thus
there exists a homomorphism $h_{a,b}=h_i$ 
that maps $a$ to the singleton $\{b\}$; so, we have
$\bigcup_{j \in [n]} h_j(a) = \bigcup_{i \in [n], |h_i(a)| = 1} h_i(a)$.
Since $g$ is a homomorphism from $\rela$ to $\unionsing(\pow(\relb)^{n})$,
we can compose $g$ and a homomorphism from $\unionsing(\pow(\relb)^{n})$ to $\relb$, 
which we know to exist by assumption,
to get a homomorphism from $\rela$ to $\relb$.
Consequently, singleton arc consistency solves $\csp(\relb)$.
\end{pf}

\section{Strength Comparison}
\label{sect:comparison}

In this section, we investigate relationships among the sets of
structures solvable by the various algorithms presented.
We show that for the structures having all constants, AC solves
a strictly smaller set of structures than LAAC does;
on the other hand, we show that there is a structure (not having all constants)
solvable by AC but not LAAC.
We then show that the structures solvable by AC or LAAC
are strictly contained in those solvable by PAC; and, 
in turn, that the structures solvable by PAC 
are strictly contained in those solvable by SAC.
We also show that 
the structures solvable by SAC 
(and hence, those solvable by any of the studied algorithms) 
all fall into the class of structures
having \emph{bounded width}; bounded width is a well-studied
condition admitting multiple 
characterizations~\cite{FederVardi,LaroseZadori,bkbw}.

\begin{prop}
\label{prop:all-constants-ac-implies-laac}
Suppose that $\relb$ is a structure having all constants.
If $\csp(\relb)$ is solvable by AC, then it is solvable by LAAC.
\end{prop}

\begin{pf}
By Theorem~\ref{thm:ac-solves}, there is a homomorphism
$f: \pow(\relb) \rightarrow \relb$.  Since the structure $\relb$
has all constants, for each $b \in B$ there is a relation symbol $R_b$
with $R_b^{\relb} = \{ (b) \}$.  
Since $(\{ b \}) \in R_b^{\pow(\relb)}$, 
it must hold that $f( \{ b \} ) \in R_b^{\relb}$, from which it follows that
$f( \{ b \} ) = b$.  The mapping $l$ defined by
$l(S, b) = f(S)$ is then a homomorphism of the type described
in Theorem~\ref{thm:laac-solves}.
\end{pf}

\begin{prop}
There exists a structure $\relb$ having all constants such that $\csp(\relb)$
is solvable by LAAC but not by AC.
\end{prop}

\begin{pf}
Take $\relb$ to be the relational structure 
with universe $\{ 0, 1 \}$ over signature
$\{ U_0, U_1, R_{(0, 0)}, R_{(1, 1)} \}$
where
$$U_0^\relb = \{ 0 \}$$
$$U_1^\relb = \{ 1 \}$$
$$R_{(0, 0)}^\relb = \{ 0, 1 \}^2 \setminus \{ (0, 0) \}$$
$$R_{(1, 1)}^\relb = \{ 0, 1 \}^2 \setminus \{ (1, 1) \}.$$
It is straightforward to verify that the mapping
$l$ defined by 
$l(\{ 0, 1 \}, b') = b'$, 
$l(\{ 0 \}, b') = 0$, and $l(\{ 1 \}, b') = 1$
for all $b' \in \{ 0, 1 \}$
is a homomorphism from $\pow(\relb) \times \relb$ to $\relb$
satisfying the condition of Theorem~\ref{thm:laac-solves}.
Hence, the problem $\csp(\relb)$ is solvable by LAAC.

To show that the problem $\csp(\relb)$ is not solvable by AC,
let $f$ be an arbitrary mapping from $\pow(B) \setminus \{ \emptyset \}$
to $B$.  We show that $f$ cannot be a homomorphism from $\pow(\relb)$
to $\relb$, which suffices by Theorem~\ref{thm:ac-solves}.
Let $b = f(\{ 0, 1 \})$.  
It holds that
$( \{ 0, 1 \}, \{ 0, 1 \} ) \in R_{(b, b)}^{\pow(\relb)}$,
but $(f(\{ 0, 1 \}), f(\{ 0, 1 \})) = (b, b) \notin R_{(b, b)}^{\relb}$,
and we are done.
\end{pf}

\begin{prop}
There exists a structure $\relb$ (not having all constants) such that $\csp(\relb)$
is solvable by AC but not by LAAC.
\end{prop}

\begin{pf}
Take $\relb$ to be the relational structure with universe $\{ 0, 1 \}$
over signature $\{ R, S \}$
where $R^{\relb} = \{ 0, 1 \}^3 \setminus \{ (0, 1, 1) \}$
and $S^{\relb} = \{ 0, 1 \}^3 \setminus \{ (1, 0, 0) \}$.
The mapping $p$ that sends each element of $\pow(B) \setminus \{ \emptyset \}$
to $0$ is a homomorphism from $\pow(\relb)$ to $\relb$,
and hence AC solves $\csp(\relb)$ by Theorem~\ref{thm:ac-solves}.

To show that the problem $\csp(\relb)$ is not solvable by LAAC,
let $f$ be an arbitrary mapping from $(\pow(B) \setminus \{ \emptyset \}) \times B$
to $B$
that satisfies
$f(\{ b \}, b') = b$ for all $b, b' \in B$.
We show that $f$ cannot be a homomorphism from
$\pow(\relb) \times \relb$ to $\relb$,
which suffices by Theorem~\ref{thm:laac-solves}.
We consider two cases depending on the value of $f( \{ 0, 1 \}, 0)$.

\begin{itemize}

\item If $f( \{ 0, 1 \}, 0) = 1$, then 
we use the facts that $(\{ 0 \}, \{ 0, 1 \}, \{ 0, 1 \}) \in R^{\pow(\relb)}$
and that $(0, 0, 0) \in R^{\relb}$;
we have that
$(f( \{0 \}, 0), f( \{ 0, 1 \}, 0), f( \{ 0, 1 \}, 0)) = (0, 1, 1)$,
which is not contained in $R^{\relb}$,
implying that $f$ is not a homomorphism of the desired type.

\item If $f( \{ 0, 1 \}, 0) = 0$, then
we use the facts that $(\{ 1 \}, \{ 0, 1 \}, \{ 0, 1 \}) \in S^{\pow(\relb)}$
and that $(0, 0, 0) \in S^{\relb}$;
we have that
$(f( \{ 1 \}, 0), f( \{ 0, 1 \}, 0), f( \{ 0, 1 \}, 0)) = (1, 0, 0)$,
which is not contained in $S^{\relb}$,
implying that $f$ is not a homomorphism of the desired type.

\end{itemize}
\end{pf}

We now proceed to study PAC, and in particular, show that the structures
solvable by AC or LAAC are solvable by PAC.

\begin{prop}
\label{prop:ac-in-pac}
Let $\relb$ be a structure.
If $\csp(\relb)$ is solvable by AC, then it is also solvable by PAC.
\end{prop}

Proposition~\ref{prop:ac-in-pac} follows directly from the algebraic
characterizations given in 
Theorems~\ref{thm:ac-solves}
and~\ref{thm:pac-solves}; 
it can also be seen to follow from
the corresponding algorithm descriptions.

\begin{theorem}
Let $\relb$ be a structure.
If $\csp(\relb)$ is solvable by LAAC, then it is also solvable by PAC.
\end{theorem}

\begin{pf}
Suppose that look-ahead arc consistency solves CSP($\relb$). 
By Theorem \ref{thm:laac-solves} there exists a homomorphism 
$l\colon \pow(\relb) \times \relb \to \relb$ such that
$l( \{ b \}, b') = b$ for all $b, b' \in B$.
We want to show that peek arc consistency solves CSP($\relb$) by using Theorem~\ref{thm:pac-solves}.
Thus, we have to show that
for all $n \geq 1$
there is a homomorphism $g_n\colon \sing(\pow(\relb)^n) \rightarrow \relb$.

Let $n\geq 1$.
Let us consider the mapping $g_n$ with $$g_n(S_1,\ldots ,S_n)=l(S_1,l(S_2,\ldots l(S_{n-1},l(S_n,b))\ldots ))$$
defined for all tuples $T=(S_1,\ldots ,S_n)\in \sing(\pow(B)^n)$ and all $b\in B$.
First we want to show that $g_n$ is well defined.
Let $b_1,b_2\in B$ with $b_1\not = b_2$, let $(S_1,\ldots ,S_n)\in \sing(\pow(B)^n)$ and let 
$i\in[n]$ be an index such that $S_i$ is a singleton. Let $S_i=\{b_*\}$ for a $b_*\in B$.
We obtain that
\begin{align*}
&    l(S_1,\ldots l(S_{i-1},l(S_{i},\ldots l(S_{n-1},l(S_n,b_1))\ldots ))\ldots )\\
=\;&   l(S_1,\ldots l(S_{i-1},l(S_{i},b'))\ldots )\\
=\;&   l(S_1,\ldots l(S_{i-1},b_*)\ldots )
\intertext{with $b'=l(S_{i+1},\ldots l(S_{n-1},l(S_n,b_1))\ldots )\in B$, because $l$ is applied to
the singleton $S_i=\{b_*\}$ and $b'$. 
Similarly, we obtain that}
&    l(S_1,\ldots l(S_{i-1},l(S_{i},\ldots l(S_{n-1},l(S_n,b_2))\ldots ))\ldots )\\
=\;&   l(S_1,\ldots l(S_{i-1},b_*)\ldots )
\end{align*} 
Consequently, $g_n$ is well defined.
Next, we  prove that $g_n$ is a homomorphism.
Let $R^{\sing(\pow(\relb)^n)}$ be a $k$-ary relation and let $(T^1,\ldots ,T^k)$ be a tuple in this relation.
Denote $T^i=(S^i_1,S^i_2,\ldots ,S^i_n)$ for all $i\in[k]$; then,
$S'_j=(S^1_j,\ldots ,S^k_j)$ has to be in $R^{\pow(\relb)}$
for all $j\in[n]$. 
Further, we know that there exists a tuple $\bar{b}=(b_1,\ldots ,b_k)\in R^\relb$, because $R^{\pow(\relb)}$ is not empty.
Since $l$ is a homomorphism, the tuple
$$g_n(S'_1, S'_2, \ldots, S'_n)=l(S'_1, l(S'_2, \ldots
l(S'_{n-1},l(S'_n,\bar{b}))\ldots ))$$
is in $R^{\relb}$. Thus, $g_n$ is a homomorphism from $\sing(\pow(\relb)^{n})$ to $\relb$.
\end{pf}

\begin{theorem}
There exists a structure $\relb$ having all constants
such that $\csp(\relb)$ is solvable by PAC but not by LAAC nor AC.
\end{theorem}

\begin{pf}
Let us consider the structure with universe $\{0,1,2\}$ over the signature $\{U_0,U_1,U_2,R_1,R_2\}$ where
$$U_0^\relb=\{(0)\}$$
$$U_1^\relb=\{(1)\}$$
$$U_2^\relb=\{(2)\}$$
$$R_1^\relb= \big(\{0,1\}\times\{0,1,2\}\big)  \setminus  \{(0,0)\}$$
$$R_2^\relb=\{(0,0),(1,2),(2,1)\}.$$
First we show that there is no homomorphism $l: \pow(\relb)\times\relb\to\relb$ such that $l(\{b\},b')=b$ for all $ b,b'$.
Let us assume there is one. 
Since $(\{0\},\{1,2\})\in R_1^{\pow(\relb)}$ and $(1,0)\in R_1^\relb$ the tuple
$( l(\{0\},1),l(\{1,2\},0))$, which is equal to $(0,l(\{1,2\},0))$,
has to be contained in $R_1^\relb$.
Thus,  $l(\{1,2\},0)$ cannot be equal to $0$.
On the other hand, $(\{1,2\},\{1,2\})\in R_2^{\pow(\relb)}$ and  $(0,0)\in R_2^\relb$
implies that $(l(\{1,2\},0),l(\{1,2\},0))$ is in $R_2^\relb$.
Therefore, $l(\{1,2\},0)$ has to be $0$, which is a contradiction.
This establishes that the structure is not solvable by LAAC;
by Proposition~\ref{prop:all-constants-ac-implies-laac},
it follows that the structure is not solvable by AC.

Next we show that for all $n$, there exists a homomorphism $f$ from $\sing(\pow(\relb)^n)$ to $\relb$.
Let $n$ be arbitrary and  let $(S_1,\dots,S_n)$ be an arbitrary $n$-tuple of $\sing(\pow(\relb)^n)$.
Further, let $i$ be the minimal number such that $S_i$ is $\{1\}$, $ \{2\}$, $ \{0,1\}$ or $\{0,2\}$;
if such an $S_i$ does not exists, then $i=0$.
The homomorphism $f$ can be defined as follows:
$$f(S_1,\dots,S_n)=
\begin{cases}
1 & \text{if  $i>0$ and $S_i$ is $\{1\}$ or $\{0,1\}$}\\
2 & \text{if  $i>0$ and $S_i$ is $\{2\}$ or $\{0,2\}$}\\
0 & \text{otherwise.}\\
\end{cases}$$
Let us verify that $f$ is indeed a homomorphism:
First of all, it is easy to see that $f(S_1,\dots,S_n)$ is in $U_i^{\relb}$
whenever $(S_1,\dots,S_n)$ is in $U_i^{\sing(\pow(\relb)^n)}$.
Next, let us consider $R_2$.
Let $(S_1,\dots,S_n)$ and $(T_1,\dots,T_n)$ be arbitrary $n$-tuples of $\sing(\pow(\relb)^n)$
such that $(S_l,T_l)$ is in $R_2^{\pow(\relb)}$ for all $l$.
Let $i$ be the minimal number such that $S_i$ is $\{1\}$,  $\{2\}$,  $\{0,1\}$ or $\{0,2\}$, and
let $j $ be the minimal number such that $T_j$ is $\{1\}$,  $\{2\}$,  $\{0,1\}$ or $\{0,2\}$,
and if such an $S_i $ or $T_j$ does not exists, then $i=0$ or $j=0$ respectively.
If $i > 0$, then $T_i$ has to be $\{1\}$,  $\{2\}$,  $\{0,1\}$ or $\{0,2\}$ and hence $0<j \leq i$.
Symmetrically, if $j > 0$, then $0<i \leq j$. Therefore, $i = j$.
Now, if  $i = j = 0$, then $(f(S_1,\dots,S_n), f(T_1,\dots,T_n)) = (0,0)$,
which is in $R_2^{\relb}$;
if $i = j > 0$, then $(f(S_1,\dots,S_n), f(T_1,\dots,T_n)) \in R_2^{\relb}$ follows directly from $(S_i,T_i)$ being in $R_2^{\pow(\relb)}$.
Finally, let us consider two arbitrary $n$-tuples $(S_1,\dots,S_n)$ and $(T_1,\dots,T_n)$ of $\sing(\pow(\relb)^n)$
such that $(S_l,T_l)$ is in $R_1^{\pow(\relb)}$ for all $l$.
If $f(S_1,\dots,S_n)=2$, then $S_i=\{2\}$ or $\{0,2\}$ and $(S_i,T_i)$ cannot be in $R_1^{\pow(\relb)}$.
If $f(S_1,\dots,S_n)=1$, then $(f(S_1,\dots,S_n), f(T_1,\dots,T_n) )$ is in $\{1\}\times\{0,1,2 \}$ and, thus, in $R_1^{\relb}$.
If $j=0$, then let $k$ be an index such that $T_k=\{0\}$.
Such an index has to exist, because $(T_1,\dots,T_n)$ is a tuple of $\sing(\pow(\relb)^n)$.
Since $(S_k,T_k)$ is in $R_1^{\pow(\relb)}$, $S_k$ has to be $\{1\}$,
and hence $f(S_1,\dots.,S_n) \in\{1,2\}$, and we appeal to one of the first two cases.
The remaining case is $i = 0$ and $j>0$.
In this case, $(f(S_1,\dots,S_n), f(T_1,\dots,T_n) )$ is in $\{0\}\times\{1,2 \}$ and therefore in $R_1^{\relb}$.
\end{pf}

We now move on to study SAC; we show that SAC is strictly more powerful
than PAC.

\begin{prop}
\label{prop:pac-in-sac}
Let $\relb$ be a structure.
If $\csp(\relb)$ is solvable by PAC, then it is also solvable by SAC.
\end{prop}

Proposition~\ref{prop:pac-in-sac} follows directly from the algebraic
characterizations given in 
Theorems~\ref{thm:pac-solves}
and~\ref{thm:sac-solves};
it can also be seen to follow from
the corresponding algorithm descriptions.

\begin{theorem}
There exists a structure $\relb$ having all constants
such that $\csp(\relb)$ is solvable by SAC but not by PAC.
\end{theorem}

\newcommand{\op}{*}

\begin{pf}
We will consider a structure that has as a polymorphism the 
idempotent binary commutative
operation $\op$ defined on the set $\{ 0, 1, 2, 3 \}$ by
$1 \op 2 = 2$, $2 \op 3 = 3$, $3 \op 1 = 1$,
and
$0 \op a = a$ for all $a \in \{ 1, 2, 3 \}$.
We consider the structure $\relb$ with universe $\{ 0, 1, 2, 3 \}$
over the signature $\{U_0, U_1, U_2, U_3, R_1, R_2 \}$
where we have
$$U_0^{\relb} = \{ (0) \}$$
$$U_1^{\relb} = \{ (1) \}$$
$$U_2^{\relb} = \{ (2) \}$$
$$U_3^{\relb} = \{ (3) \}.$$
$$R_1^{\relb} = \{ 0, 1, 2, 3 \}^2 \setminus \{ (0, 0) \},$$
$$R_2^{\relb} = \{ (1, 2), (2, 3), (3, 1), (0, 0) \}$$
It is straightforward to verify that this structure $\relb$ has
the operation $\op$ as a polymorphism.
The solvability of $\relb$ follows from Theorem~\ref{thm:twosem-sac},
which is proved in the next section; see also
the discussion in Example~\ref{ex:twosem}.

To show that peek arc consistency does not solve $\csp(\relb)$,
we prove that there is no homomorphism from 
$\sing(\pow(\relb)^2)$ to $\relb$, which is sufficient by 
Theorem~\ref{thm:pac-solves}.
Define $\tup{t_1} = ( \{ 0 \}, \{ 1, 2, 3 \} )$ and
$\tup{t_2} = (\{ 1, 2, 3 \}, \{ 0 \} )$.
It is straightforward to verify that 
$(\tup{t_1}, \tup{t_2}) \in R_1^{\pow(\relb)^2}$;
since each of the tuples $\tup{t_1}, \tup{t_2}$ contains a singleton,
it holds that
$(\tup{t_1}, \tup{t_2}) \in R_1^{\sing(\pow(\relb)^2)}$.
Assume, for a contradiction, that $h$ is a homomorphism from
$\sing(\pow(\relb)^2)$ to $\relb$.
It then holds that
$(h(\tup{t_1}), h(\tup{t_2})) \in R_1^{\relb}$.
Since $(0, 0) \notin R_1^{\relb}$, we have that 
one of the values 
$h(\tup{t_1}), h(\tup{t_2})$ is not equal to $0$.
Let us assume that $h(\tup{t_1})$ is not equal to $0$;
the other case is symmetric.  Denote $h(\tup{t_1})$ by $b$;
we have $b \in \{ 1, 2, 3 \}$.
Since each of the two tuples
$( \{ 0 \}, \{ 0 \} )$, $( \{ 1, 2, 3 \}, \{ 1, 2, 3 \} )$
is contained in $R_2^{\pow(\relb)}$, we have that
$(( \{ 0 \}, \{ 1, 2, 3 \} ), ( \{ 0 \}, \{ 1, 2, 3 \} ))
\in R_2^{\sing(\pow(\relb)^2)}$.
It follows that $( b, b ) \in R_2^{\relb}$, but since no tuple of the form
$(c, c)$ with $c \in \{ 1, 2, 3 \}$ is contained in $R_2^{\relb}$,
we have reached our contradiction.
\end{pf}

\newcommand{\dom}{\mathrm{Dom}}

We close this section by showing that the structures solvable
by SAC, and hence those solvable by any of the algorithms studied here,
fall into the class of structures having \emph{bounded width}.
We begin by defining bounded width.
A {\em partial homomorphism} from $\rela$ to $\relb$
is a mapping $f:A'\rightarrow B$, where $A'\subseteq A$,
that defines a homomorphism
to $\relb$
from the substructure of $\rela$ induced by $A'$.
When $f$ and $g$ are partial homomorphisms we say that $g$ {\em extends} $f$,
denoted by $f\subseteq g$, 
if $\dom(f)\subseteq\dom(g)$ and $f(a)=g(a)$ for every $a\in\dom(f)$.

\begin{definition}
Let $k>1$. A {\em $k$-strategy} for an instance $(\rela,\relb)$ is a nonempty collection $H$ of partial homomorphisms from $\rela$ to $\relb$
satisfying the following conditions:
\begin{enumerate}
\item (restriction condition) if $f\in H$ and $g\subseteq f$, then $g\in H$;
\item (extension condition) if $f\in H$, $|\dom(f)|<k$, and $a\in A$, there is $g\in H$ such that $f\subseteq g$
and $a\in\dom(g)$.
\end{enumerate}
\end{definition}

When $H$ is a $k$-strategy for $(\rela,\relb)$ and 
$a_1,\dots,a_j\in A$ is a sequence, 
we define $H_{a_1,\dots,a_j}\subseteq B^j$ to be the relation
$$\{(f(a_1),\dots,f(a_j)) ~|~ f\in H, \dom(f)=\{a_1,\dots,a_j\}\}.$$

\begin{definition}
Let $\relb$ be a structure and $k \geq 1$. 
We say that $\csp(\relb)$ has {\em width} $k$
if for all structures $\rela$ the following holds:
if there is a $(k+1)$-strategy for $(\rela,\relb)$ then there is a homomorphism $\rela\rightarrow\relb$.
We say that $\csp(\relb)$ has {\em bounded width} if it has width $k$ 
for some $k \geq 1$.
\end{definition}

\begin{prop}
Let $\relb$ be a structure.
If $\csp(\relb)$ is solvable by SAC, then $\csp(\relb)$ has bounded width.
\end{prop}

\begin{pf}
Let
$r$ be the maximum of all the arities of the signature of $\relb$,
and set $k = \max(2, r+1)$.
We shall show that for any instance $\rela$ of $\csp(\relb)$,
if $H$ is a 
$k$-strategy for $(\rela,\relb)$, then the instance $(\rela,\relb)$ has the SACC, which suffices.

Let us define the mapping $s:A\rightarrow \pow(B)\setminus\{ \emptyset \}$ as $s(a)=H_a$. Furthermore, for every $a\in A$, $b\in s(a)$,
define $h_{a,b}: A \rightarrow \pow(B)\setminus\{\emptyset\}$ as the mapping $h_{a,b}(a')=\{b' ~|~ (b,b')\in H_{a,a'} \}$. Note that the extension property of $H$
guarantees that, for every $a'\in A$,  $h_{a,b}(a')$ is, indeed, nonempty. It follows from the definition of $h_{a,b}$ that $h_{a,b}(a)=\{b\}$, and that for 
all $a' \in A$, $h_{a,b}(a') \subseteq s(a')$. 

It is only necessary to show that $h_{a,b}$ defines a homomorphism from $\rela$ to $\pow(\relb)$.  Let $R^{\rela}$ be any relation in $\rela$, let
$(a_1,\dots,a_i)\in R^{\rela}$, 
and let $S_j=h_{a,b}(a_j)$ for each $j \in [i]$.
In order to prove that $(S_1,\dots,S_i)\in R^{\pow(\relb)}$ it suffices to show
that for every $j\in [i]$ and every $b_j\in S_j$, there exists some $(c_1,\dots,c_i)\in R^{\relb} \cap (S_1\times\cdots\times S_i)$ with $c_j=b_j$. 
This is a direct consequence of the properties of the strategy. Indeed, by the definition of $h_{a,b}$ we know that $(b,b_j)\in H_{a,a_j}$ and then, by an iterative
application of the extension property, we can show that there exists an extension $(b,c_1,\dots,c_i)\in H_{a,a_1,\dots,a_i}$ with $c_j=b_j$. The fact that $H$ contains
only partial homomomorphisms guarantees that $(c_1,\dots,c_i)\in R^\relb$. Finally, it follows from the restriction condition that for every $l \in [i]$, we have
$c_l\in S_{l}$.
\end{pf}

\section{Tractability via singleton arc consistency}
\label{sect:tractability}

\subsection{Majority operations}

An operation $m\colon B^3\to B$ is a \emph{majority} operation
if it satisfies the identity $m(x,y,y)=m(y,x,y)=m(y,y,x)=y$ for all $x,y\in B$.
Relative to a majority operation $m \colon B^3 \to B$,
when $I \subseteq J \subseteq B$, 
we say that $I$ is an \emph{ideal} of $J$ if for every $x,y,z\in J$ such that $x,z\in I$ we have $m(x,y,z)\in I$.  We will establish the following result.

\begin{theorem}
\label{the:SACCmajority}
If $\relb$ is a structure that has a majority polymorphism, then
singleton arc consistency solves $\csp(\relb)$.
\end{theorem}

The proof is obtained by using a strengthened version of the Prague
strategy defined by Barto and Kozik~\cite{bkbw}.

In this section, for the sake of readability, we will typically use the notation
$t[i]$ to denote the $i$th entry of a tuple $t$.

We introduce the following definitions relative to an instance $(\rela, \relb)$
with signature $\sigma$.
A \emph{pattern} $p$ of $\rela$ is a sequence $a_1,e_1,a_2,\dots,e_{m-1},a_m$ such that $a_1,\dots,a_m$ are elements of $A$
and for every $n \in [m-1]$, 
we have that
$e_n$ is a triple $(R,i,j)$ where $R$ is a symbol in $\sigma$ and 
$i,j$ are indices such that there is a tuple 
$t\in R^{\rela}$ with $t[i]=a_n$ and $t[j]=a_{n+1}$. The {\em length} of pattern $p$ is defined to be $m$.
A pattern is a \emph{cycle} if $a_1=a_m$.
By a \emph{set system}, we mean any mapping 
$H: A \rightarrow \pow(B) \setminus \{ \emptyset \}$.

A pattern
$q = b_1,e'_1,\dots,e'_{m-1},b_m$ 
of $\relb$ having the same length as a pattern $p$ of $\rela$
is a {\em realization} of $p$ if $e_n=e'_n$ for all $n \in [m-1]$. 
The pair $(b_1,b_m)$ is said to be a {\em support} of $p$.
For a set system $H$,
if it holds that $b_n\in H(a_n)$ for all $n\in [m]$ then 
$(b_1,b_m)$ is said to be a support of $p$ {\em inside} $H$. 

A set system $H$ is a \emph{weak strategy}
if for every pattern $p=a_1,e_1,\dots,e_{m-1},a_m$ of $\rela$,
and every $b_1\in H(a_1)$ there exists some $b_m\in H(a_m)$ such that $(b_1,b_m)$ supports $p$ inside $H$.
A set system $H$ is a \emph{strong strategy} if
for every cycle $p=(a=a_1,\dots,a_m=a)$ in $\rela$ and every $b\in H(a)$, 
the pair $(b,b)$ supports $p$ inside $H$.
Note that every strong strategy is a weak strategy and that the class of weak strategies
remains the same if, in the definition of weak strategy,  
one replaces
``every pattern $p=a_1,e_1,\dots,e_{m-1},a_m$''
by ``every pattern $p=a_1,e_1,\dots,e_{m-1},a_m$ of length $m=2$''.

\begin{observation}
\label{obs:arcconsistency}
Every strong strategy is a weak strategy,
relative to an instance $(\rela, \relb)$.
\end{observation}
\begin{pf}
For a pattern
$p=a_1,e_1,\dots,e_{m-1},a_m$ of ${\mathbf A}$,
one needs only to apply the definition of strong strategy to the the pattern $a_1,e_1,\dots,e_{m-1},a_m,e_{m-1}^{-1},a_{m-1},\dots,e_1^{-1},a_1$, where $(R,i,j)^{-1}$ is defined to be $(R,j,i)$.
\end{pf}

\begin{lemma}
\label{le:SACCtostrong}
There exists a strong strategy
for an instance $(\rela, \relb)$ having the SACC.
\end{lemma}

\begin{pf}
Let $s: A \rightarrow \pow(B) \setminus \{ \emptyset \}$, $\{ h_{a, b} \}$
be the mappings witnessing that $(\rela,\relb)$ has the SACC.
We claim that the set system $H$ defined
by $H(a)=s(a)$ for all $a\in A$ is a strong strategy. Indeed, let $p=a_1,e_1,\dots,a_m$ be a pattern of $A$ with $a_1=a_m=a$ and let $b\in H(a_1)$.
We claim that there exists a realization $b_1,e_1,\dots,b_m$ of $p$ with $b_1=b_m=b$ such that for every $1\leq n\leq m$, $b_n\in h_{a,b}(a_n)$.
The realization is constructed in an inductive manner. First, set $b_1$ to $b$. Assume now
that $b_{n-1}$ is already set and let $e_{n-1}$ be $(R,i,j)$.
There exists a tuple $(x_1,\dots,x_r)\in R^{\rela}$ such that $x_i=a_{n-1}$ and $x_j=a_n$.
Since $h_{a,b}$ is a homomorphism,
the subset $S\subseteq B^r$ defined by $\pi_l S=h_{a,b}(x_l)$ for every $1\leq l\leq r$ is a subset of $R^{\relb}$.
From $b_{n-1}\in h_{a,b}(x_i)$ it follows that there exists a tuple $(y_1,\dots,y_r)\in S$ with $y_i=b_{n-1}$. Define $b_n$ to be $y_j$.
Since, by definition of SACC strategy
$h_{a,b}(a)=\{b\}$, it follows that $b_m=b$.
\end{pf}

We now prove the following lemma, which, as we explain after the proof,
essentially establishes the desired theorem.  In the course of proving
this lemma, we establish a number of observations.

By a minimal strong strategy, we mean minimal with respect to the ordering
where for two strategies $H, H'$, we consider $H \subseteq H'$
if $H(a) \subseteq H'(a)$ for all $a \in A$.

\begin{lemma}
\label{le:singletonstrong}
If the relations of $\relb$ are invariant under a majority operation $\phi$
and $H$ is a minimal strong strategy then 
for every $a \in A$, the set $H(a)$ is a singleton.
\end{lemma}

\begin{pf}
Towards a contradiction assume that 
$H$ is a minimal strong strategy and $a^* \in A$ is such that
$H(a^*)$, is not a singleton. 
Consider the digraph $G$ whose nodes are of the form $(a,C)$ with
$a\in A$ and $C\subseteq H(a)$, and there is an edge from $(a,C)$ to $(a',C')$
if there is a pattern $p=a_1,\dots,a_m$ with $a=a_1$ and $a'=a_m $ in ${\mathbf A}$ such that the following holds:
$C'$ is the set containing all $b'\in H(a')$ such that $(b,b')$ is supported by $p$ inside $H$ for some $b\in C$.

\begin{observation}
\label{observation3}
Let $p=a_1,e_1,\dots,a_m$ be a pattern, let $1<i<m$, let $q$ be the pattern $a_1,e_1,\dots,a_i$
and $r$ be the pattern $a_i,e_i\dots,e_m$. If $q$ defines an edge from $(a_1,C_1)$ to $(a_i,C_i)$
and $r$ defines an edge from $(a_i,C_i)$ to $(a_m,C_m)$ then $p$ defines an edge from $(a_1,C_1)$ to $(a_m,C_m)$.
Hence, the graph $G$ is transitive.
\end{observation}

The following observation follows from the definition of strong strategy.

\begin{observation}\label{observation1}
If there is an edge from $(a,C)$
to $(a,C')$ in 
$G$, 
then necessarily $C\subseteq C'$.
\end{observation}

\begin{observation}\label{observation2}
If there is an edge from $(a,C)$
to $(a',C')$ in 
$G$, and $C$ is an ideal of $H(a)$, then $C'$ is an ideal of $H(a')$.
\end{observation}

\begin{pf} [ (Observation~\ref{observation2})]
Let us prove the claim by induction on the length $m$ of the pattern that defines the edge.

Assume first that $m=2$. Let $a,(R,i,j),a'$ be any such pattern.
Let $x_1,x_2,x_3\in H(a')$ and assume that two of them, say $x_1,x_3$, belong to $C'$. It follows, by the definition of edge,
that for every $n\in\{1,3\}$ there exists tuple $t_n\in R^{\relb}$ with $t_n[j]=x_n$ and $t_n[i]\in C$.
Also, it follows by considering pattern $a',(R,j,i),a$ and from the fact that $H$ is a weak strategy that there exists a tuple 
$t_2\in R^{\relb}$ with $t_2[i]\in H(a)$ and $t_2[j]=x_2$.
Consider now tuple $t=\phi(t_1,t_2,t_3)$. Since $C$ is an ideal of $H(a)$ we have that $t[i]\in C$. Hence, we conclude that $\phi(x_1,x_2,x_3)=t[j]\in C'$.

The case $m>2$ follows from the inductive hypothesis and Observation~\ref{observation3}.
\end{pf}

Now, let $G'$ be the subgraph of $G$ induced by all nodes $(a,C)$ such that $C$ is an ideal of $H(a)$ and $C\neq H(a)$.
Observe that as $H(a^*)$ is not a singleton, the graph $G'$ is nonempty, because every singleton is an ideal.

A subset $M$ of vertices of a directed graph is a \emph{strongly connected component}
if for every pair $(v,w)\in M^2$ there exists a path from $v$ to $w$ consisting only of vertices in $M$. It is a \emph{maximal strongly connected component} if additionally,
there is no edge $(v,w)$ with $v\in M$ and $w\not\in M$.

Let $M$ be a maximal strongly connected component of $G'$. 
The following observation is a direct consequence of Observations~\ref{observation3} and \ref{observation1}.

\begin{observation}\label{observation4}
The maximal strongly connected component $M$ cannot have
two vertices $(a,C)$, $(a,C')$ with $C\neq C'$.
\end{observation}

We shall construct a new strong strategy $H'$ as follows. If $(a,C)$ belongs to $M$, then
set $H'(a)=C$ otherwise set $H'(a)=H(a)$. Clearly $H'$ is strictly smaller than $H$.

We shall start by showing that $H'$ is a weak strategy. By the note following the definition of weak strategy
it is only necessary to show that for every pattern $p=a_1,e_1,a_2$ of length $2$ of $\rela$ and
every $b_1\in H'(a_1)$,  there exists a support $(b_1,b_2)$ of $p$ inside $H'$.

We do a case analysis.
If $(a_2,H'(a_2))$ does not belong to $M$ the claim follows from the fact that $H$ is a weak strategy.
Assume now that $(a_2,H'(a_2))$ belongs to $M$. Consider the pattern $p=a_2,e_1^{-1},a_1$ where $(R,i,j)^{-1}=(R,j,i)$. 
This pattern defines an edge (in $G$) from $(a_2,H'(a_2))$
to a node $(a_1,C)$. Observe, that by the definition of the edges of $G$, we know that for every element $b\in C$ there
is some $b'\in H'(a_2)$ such that $(b,b')$ is supported by $p$ inside $H$. Hence we only need to show that $H'(a_1)\subseteq C$.

If $(a_1,C)$ is in $G'$ then, since $M$ is a maximal strongly connected component of $G'$,
we have that $(a_1,C)$ belongs to $M$ as well and hence $C=H'(a_1)$.
If $(a_1,C)$ is not in $G'$ this must be because $C$ is not an ideal of $H(a_1)$ or because $C=H(a_1)$. We can rule out the
first possibility in the following way: by the definitions of $G'$ and $H'$, $H'(a_2)$ is an ideal of $H(a_2)$. It follows
by observation~\ref{observation2} that $C$ is an ideal of $H(a_1)$. In consequence $C=H(a_1)$ and the proof that $H'$ is
a weak strategy is concluded.

It remains to show that $H'$ is a strong strategy. Let $p=a_1,e_1,\dots,e_{m-1},a_m$ be any cycle in $\mathbf A$ with $a_1=a_m=a$ and let $b$ be any element in $H'(a)$.
Since $H'$ is a weak strategy we know that there is a realization $b_1,\dots,b_m$ of $p$ with $b_1=b$ inside $H'$. 
Notice that we do not necessarily have $b_m = b$.
Symmetrically, by considering pattern $a_m,e^{-1}_{m-1}, \ldots, e^{-1}_1,a_1$ we know that there is a realization 
$d_m,e^{-1}_{m-1}, \ldots, e^{-1}_1,d_1$ of $p$ with $d_m=b$ inside $H'$. 
Also, since $H$ is a strong strategy
we know that there exists a realization $c_1,e_1,\dots,c_m$ of $p$ such that $c_1=c_m=b$ inside $H$ 
(but not necessarily inside $H'$). Finally
consider the sequence $x_1,\dots,x_m$ defined by $x_j=\phi(b_j,c_j,d_j), 1\leq j\leq m$. This sequence is a realization of $p$. 
Furthermore, we have that $x_1=x_m=b$.
It remains to show that it is inside $H'$. Indeed, for every $1\leq j\leq m$, $\{b_j,d_j\}\subseteq H'(a_j)$ and
$c_j\in H(a_j)$. Since $H'(a_j)$ is an ideal of $H(a_j)$ the claim follows.\end{pf}

\begin{pf}
(Theorem~\ref{the:SACCmajority}) 
Suppose that the instance $(\rela, \relb)$
has the SACC and that $\relb$ has the majority polymorphism $\phi$.
By Lemmas~\ref{le:SACCtostrong} and~\ref{le:singletonstrong} 
there exists a strong strategy $H$ 
for $(\rela, \relb)$
such that $H(a)$ is a singleton for every $a\in A$. 
Consider now the mapping $h: A\rightarrow B$
that maps every $a\in A$ to the only element in $H(a)$. 
We claim that $h$ is a homomorphism from $\rela$ to $\relb$. 
Indeed, let $R$ be any relation symbol, and 
$(a_1,\dots,a_r)$ be any tuple in $R^{\rela}$. Fix any $1\leq i,j\leq r$ and consider pattern $a_i,(R,i,j),a_j$. It follows by the definition
of strong strategy that there is a $t\in R^{\relb}$ such that $t[i]=h(i)$ and $t[j]=h(j)$.
Since $R^{\relb}$ is necessarily $2$-decomposable~\cite{CCC},
$h$ is a homomorphism.
\end{pf}


\subsection{2-semilattice operations}

\newcommand{\mul}{\star}
\newcommand{\maxscc}[1]{\overline{#1}}


\newcommand{\lsa}{\langle}
\newcommand{\rsa}{\rangle}

A 2-semilattice $\algg = (G, \mul)$ consists of a set $G$, which in this paper we assume to be finite, and a binary operation
$\mul$ satisfying
$x \mul x = x$ (idempotency), 
$x \mul y = y \mul x$ (commutativity), and 
$x \mul (x \mul y) = (x \mul x) \mul y$ (restricted associativity).

Each 2-semilattice naturally induces a directed graph 
$(G, E)$ where $(a, b) \in E$ if and only if $a \mul b = b$.
When $(a,b)\in E$, we also write $a \leq b$.
The graph $(G,E)$ is connected, since  $a \mul (a \mul b) = b \mul (a \mul b) = a \mul b$ for any $a,b\in G$, and  therefore, $a,b \leq a \mul b$.
Each 2-semilattice has a unique maximal strongly connected component, that is a strongly connected component with no outgoing edges,
denoted by $\maxscc{G}$. The component $\maxscc{G}$ is also the unique strongly connected component
of $(G, E)$ such that for any $a \in G$, there exists $b \in \maxscc{G}$
such that $a \leq b$.
In this section, we will prove 
that
a certain class
of 2-semilattices
is tractable via singleton arc consistency.
Our treatment of 2-semilattices is inspired and influenced by
the study conducted by Bulatov~\cite{twosemilattices}, who proved
that they are polynomial-time tractable.

A 2-semilattice $\algg = (G, \mul)$ is an algebra.
By an \emph{algebra}, we mean a pair $(A, O)$ consisting of a set $A$,
the \emph{universe} of the algebra, and a set $O$ of operations on $A$.
A \emph{congruence} of an algebra is an equivalence relation
preserved by the operation(s) of the algebra, and an
algebra is \emph{simple} if its only congruences are trivial
(that is, if its only congruences are the equality relation on $A$
and $A \times A$, where $A$ is the universe of the algebra).

We will begin by proving some general results on singleton arc consistency.
In the following discussion, a \emph{subalgebra}
is defined, with respect to
a relational structure $\relb$, as a subset $S \subseteq B$
that is preserved by all polymorphisms of $\relb$.
For an arbitrary subset $T \subseteq B$, we use $\lsa T \rsa$
to denote the smallest subalgebra containing $T$.

\begin{prop}
\label{prop:poly-ac}
Suppose that $g_1, \ldots, g_k: \rela \rightarrow \pow(\relb)$
are homomorphisms, and suppose that $f: B^k \rightarrow B$ 
is a polymorphism of $\relb$.
Then the map $g: A \rightarrow \pow(B) \setminus \{ \emptyset \}$
defined by $g(a) = f(g_1(a), \ldots, g_k(a))$ for all $a \in A$
is a homomorphism $\rela \rightarrow \pow(\relb)$.
\end{prop}

For an operation $f: B^k \rightarrow B$ and a sequence of subsets
$B_1, \ldots, B_k \subseteq B$, by the notation $f(B_1, \ldots, B_k)$, we denote
the set $\{ f(b_1, \ldots, b_k) ~|~ b_1 \in B_1, \ldots, b_k \in B_k \}$.
Regarding this notation, it is easy to verify that $f$ can be understood as a polymorphism of $\pow(\relb)$ if $f$ is a polymorphism of $\relb$.
Proposition~\ref{prop:poly-ac} follows straightforwardly from the definitions.

\begin{prop}
\label{prop:subalgebra-ac}
Suppose that $h: \rela \rightarrow \pow(\relb)$ is a homomorphism.
Then the map $h'$ defined by $h'(a) = \lsa h(a) \rsa$ for all $a \in A$
is also a homomorphism $\rela \rightarrow \pow(\relb)$.
\end{prop}

\begin{pf}
Repeatedly apply Proposition~\ref{prop:poly-ac}
with a polymorphism $f$ and $g_1 = \cdots = g_k = h$, each time
taking the resulting $g$ and updating $h$ to be $h \cup g$.
Note that at each step, the new $h$ is a homomorphism $\rela \rightarrow \pow(\relb)$,
since the union operation $\cup$ is a polymorphism of $\pow(\relb)$.
When no changes can be made, the resulting $h$ is the desired $h'$.
\end{pf}

Let us say that a CSP instance $(\rela, \relb)$ 
has the \emph{subalgebra SACC} if $(\rela, \relb)$ has the SACC
relative to mappings $s, \{ h_{a, b} \}$ such that
for all $a \in A$, the set $s(a)$ is a subalgebra,
and for all $a, a' \in A$, $b \in s(a)$, the set
$h_{a, b}(a')$ is a subalgebra.

\begin{prop}
\label{prop:sacc-implies-subalgebra-sacc}
If a pair $(\rela, \relb)$ of similar structures
has the SACC,
and all polymorphisms of $\relb$ are idempotent, 
then it has the subalgebra SACC.
\end{prop}

\begin{pf}
Suppose that $(\rela, \relb)$ has the SACC with respect to 
the mappings $s, \{ h_{a, b} \}$.
Set $s'(a) = \lsa s(a) \rsa$ for all $a \in A$,
and $h'_{a, b}(a') = \lsa h_{a, b}(a') \rsa$ for all 
$a, a' \in A$, $b \in s(a)$.
Clearly, for all such $a, a', b$ we have
$h'_{a, b}(a') \subseteq s'(a')$, and 
also, that $h'_{a, b}$ is a homomorphism $\rela \rightarrow \pow(\relb)$
(by Proposition~\ref{prop:subalgebra-ac}).
Let $b$ be an element in $s'(a) \setminus s(a)$ for some $a \in A$.
We need to show that there exists a homomorphism
$h'_{a, b}$ that satisfies the two conditions of
Definition~\ref{def:sacc} with respect to $s'$, and that also satisfies
the subalgebra condition.
As $s'(a)$ is defined as $\lsa s(a) \rsa$, it holds that
$s'(a) = \{ f(b_1, \ldots, b_k) ~|~ f \mbox{ a polymorphism of } \relb; b_1, \ldots, b_k \in s(a) \}$; the containment $\supseteq$ is clear by definition of subalgebra,
and the containment $\subseteq$ follows from the fact that 
the right hand side is a subalgebra, which in turn follows from the fact 
that the set of polymorphisms of $\relb$ forms a clone and
is closed under composition~\cite{szendrei86-clones}. 
Thus, there exists a polymorphism $f$ of $B$
and elements $b_1, \ldots, b_k \in s(a)$ such that
$b = f(b_1, \ldots, b_k)$.  
Let $g'_{a, b}$ be the homomorphism obtained from 
Proposition~\ref{prop:poly-ac} with $g_i = h_{a, b_i}$ and $f$.
Set $h'_{a, b}(a') = \lsa g'_{a, b}(a') \rsa$ for all $a' \in A$.
The homomorphism $h'_{a, b}$ has the desired properties.
\end{pf}

We now turn to prove our tractability result.
We will now use the term subalgebra to refer to
a \emph{subalgebra} of a 2-semilattice $(B, \mul)$, that is, a subset of $B$
preserved by $\mul$.  Note, however, that we will be working
with a relational structure $\relb$ assumed to have $\mul$ as a polymorphism,
so a subalgebra in the previous sense  
(that is, with respect to $\relb$)
will also be a subalgebra in 
this sense.
An algebra $(B, \mul)$ having a binary operation is \emph{conservative}
if for all $b, b' \in B$, it holds that $b \mul b' \in \{ b, b' \}$.
The following is the statement of our tractability result.

\begin{theorem}
\label{thm:twosem-sac}
Let $(B, \mul)$ be a conservative 2-semilattice such that every strongly connected
subalgebra is simple.
If $\relb$ is a structure having $\mul$ as a polymorphism, then 
singleton arc consistency solves $\csp(\relb)$.
\end{theorem}

\begin{example}
\label{ex:twosem}
We consider the binary operation $\op$ 
on $\{ 0, 1, 2, 3 \}$ defined by the following table.
\begin{center}
$
\begin{tabular}{|c|c|c|c|c|}
\hline
\op &  0 & 1 & 2 & 3 \\  \hline
0 &       0 & 1 & 2 & 3 \\ \hline
1 &       1 & 1 & 2  & 1   \\ \hline
2 &       2 & 2  & 2 &  3  \\ \hline
3 &       3 & 1 &  3 & 3 \\ \hline
\end{tabular}
$
\end{center}
It is straightforward to verify that this operation is commutative and
conservative, and is a 2-semilattice.
The graph induced by this operation has edges
$(0, 1), (0, 2), (0, 3), (1, 2), (2, 3), (3, 1)$,
as well as self-edges on each of the vertices.
There is thus just one strongly connected component of size strictly
greater than one, namely, the component $\{ 1, 2, 3 \}$.
This is a subalgebra of the algebra $(\{ 0, 1, 2, 3 \}, \op)$
and is readily verified to be simple.
Hence, the tractability via singleton arc consistency of any
structure preserved by the operation $\op$ follows
from Theorem~\ref{thm:twosem-sac}.
\end{example}

We will make use of the following results.
For our purposes here, a \emph{subdirect product} of algebras 
$\alga_1, \ldots, \alga_k$ is a subalgebra
$S$ of $A_1 \times \cdots \times A_k$ such that
for each $i \in [k]$, it holds that $\pi_i S = A_i$.

\begin{lemma}
\label{lemma:subdir-scc}
Suppose that $S$ is a subdirect product of 2-semilattices
$S_1, \ldots, S_n$.  Then 
$S \cap (\maxscc{S_1} \times \cdots \times \maxscc{S_n})$
is a subdirect product of $\maxscc{S_1}, \ldots, \maxscc{S_n}$.
\end{lemma}

\begin{pf}
Immediate from \cite[Lemma 3.2]{twosemilattices}.
\end{pf}

\begin{definition}\label{def:almost-trivial}
A relation $S \subseteq B^n$ is \emph{almost trivial}
if there exists a partition $I_1, \ldots, I_k$ of $[n]$
such that
\begin{itemize}

\item $t \in S$ if and only if 
for all $i \in [k]$, it holds that $\pr_{I_i} t \in \pr_{I_i} S$; and,

\item for each $j \in [k]$, it holds that 
$\pr_{I_j} S$ has the form 
$\{ (\pi_1(p), \pi_2(p), \ldots, \pi_m(p)) ~|~ p \in [q] \}$ for some $q \geq 1$ and
where each mapping $\pi_i$ is a bijection from $[q]$ to a subset of $B$.

\end{itemize}
\end{definition}

\begin{prop} 
\label{prop:subdir-of-simple-sc}
A subdirect product of simple strongly connected 2-semilattices
is an almost trivial relation, and is hence itself strongly connected.
\end{prop}

\begin{pf}
Immediate from~\cite[Proposition 3.1]{twosemilattices}.
\end{pf}

\begin{prop}
\label{prop:sacc-almost-trivial}
Let $(\rela, \relb)$ be an instance that has the SACC with respect to 
$s: A \rightarrow \pow(\relb) \setminus \{ \emptyset \}$.
If for each tuple $(a_1, \ldots, a_k) \in R^{\rela}$,
it holds that $R^{\relb} \cap (s(a_1) \times \cdots \times s(a_k))$
is almost trivial, then
there is a homomorphism from $\rela$ to $\relb$.
\end{prop}

\begin{pf}
Consider the following graph $G=(A,E)$,
where $\{a,b\}\in E$ if and only if there is a relation $R^\rela$ in $\rela$,
and, if $I_1,\dots,I_k$ is its partition regarding almost triviality of $R^{\relb} \cap (s(a_1) \times \cdots \times s(a_k))$,
there further is an $l\in [k]$ and a tuple $(a_1,\dots,a_m)\in R^{\rela}_{I_l}$ such that there are $i,j$ with $a=a_i$ and $b=a_j$.
For each connected component $C$ of the graph $G$ arbitarily choose $a\in C$ and $b\in s(a)$.
Since arc consistency can be established when $a$ is set to $b$ and using the structure of the projected relations $R^\relb_{I_l}$,
there exists a unique extension of  $a\mapsto b$ to a homomorphism on $C$.
Because of  the first property of Definition~\ref{def:almost-trivial} the homomorphisms on the single components can be combined to a homomorphism on $\rela$.
\end{pf}

The following is the main result used to prove
Theorem~\ref{thm:twosem-sac}.

\begin{theorem}
\label{thm:sacc-on-max-components}
Suppose that $\relb$ satisfies the hypotheses of
Theorem~\ref{thm:twosem-sac}, and
suppose that $(\rela, \relb)$ has the subalgebra SACC
via $s: A \rightarrow \pow(B) \setminus \{ \emptyset \}$.
Then, $(\rela, \relb)$ has the SACC via the map 
$s': A \rightarrow \pow(\relb) \setminus \{ \emptyset \}$
defined by $s'(a) = \maxscc{s(a)}$ for all $a \in A$.
\end{theorem}

\begin{pf}
Let $a \in A$ and $b \in \maxscc{s(a)}$.  
By hypothesis, there exists a homomorphism $h: \rela \rightarrow \pow(\relb)$
where $h(a) = \{ b \}$ and for all $a' \in A$, it holds that 
$h(a') \subseteq s(a')$.
We want to show that there exists a homomorphism
$h': \rela \rightarrow \pow(\relb)$ where $h'(a) = \{ b \}$
and for all $a' \in A$, it holds that $h'(a') \subseteq s'(a')$.
Define $h'(a)$ as $\maxscc{h(a)}$ if $h(a) \cap s'(a) \neq \emptyset$,
and as $s'(a)$ otherwise.
Observe that in the first case, we have 
$h'(a) = \maxscc{h(a)} \subseteq s'(a)$,
and that in both cases, the subset $h'(a)$ is a subalgebra.

We claim that $h'$ is a homomorphism from $\rela$ to $\pow(\relb)$.
Let $a \in R^{\rela}$ be a tuple in $\rela$.
For the sake of notation, we assume that
$a = (a_1, \ldots, a_{k+l})$,
$I = \{ 1, \ldots, k \}$, $J = \{ k+1, \ldots, k+l \}$,
and that $I$ contains exactly the coordinates $i \in [k+l]$
such that $h(a_i) \cap s'(a_i) \neq \emptyset$,
so that $h'(a_i) = \maxscc{h(a_i)}$ for all $i \in I$ and
$h'(a_j) = s'(a_j)$ for all $j \in J$.
Let ${T = (\pr_I R^{\relb} \cap (s(a_1) \times \cdots \times s(a_k))) \cap (\maxscc{s(a_1)} \times \cdots \times \maxscc{s(a_k)})}$.
By Lemma~\ref{lemma:subdir-scc}, we have that relation $T$ is a subdirect product of
$\maxscc{s(a_1)}, \ldots, \maxscc{s(a_k)}$.
Further, let $W = {   (R^{\relb} \cap (s(a_1) \times \cdots \times s(a_{k+l}))) \cap (\maxscc{s(a_1)} \times \cdots \times \maxscc{s(a_{k+l})})   }$.
By Lemma~\ref{lemma:subdir-scc}, we have that $W$ is a subdirect product of
$\maxscc{s(a_1)}, \ldots, \maxscc{s(a_{k+l})}$.
Clearly, $\pr_I W \subseteq T$.  
We show that $T \subseteq \pr_I W$ (and hence that
$T = \pr_I W$), as follows.
Let $t$ be a tuple in $T$.
Let $w$ be a tuple in $W$ (such a tuple can be obtained,
for instance, by $\mul$-multiplying together all tuples of 
$R^{\relb}\cap (s(a_1) \times \cdots \times s(a_{k+l}))$, in any order).
By our assumption on $\relb$ and by 
Proposition~\ref{prop:subdir-of-simple-sc}, there is 
a sequence of tuples $u_1, \ldots, u_m$ in $T$ such that
$\pr_I w \leq u_1 \leq \cdots \leq u_m = t$.
We hence have tuples $v_1, \ldots, v_m$ with $v_i \in R^{\relb}\cap (s(a_1) \times \cdots \times s(a_{k+l}))$
and $\pr_I v_i = u_i$ for each $i \in [m]$.
The product $(\cdots ((w \mul v_1) \mul v_2) \mul \cdots \mul v_m)$
gives a tuple in $W$ whose projection onto $I$ is $t$.

By Proposition~\ref{prop:subdir-of-simple-sc}, it holds that
$W$ is almost trivial with respect to a partition $\{ I_i \}$.
Remove from the $I_i$ any coordinates $l$ such that $s'(a_l)$ has just one element.
We now show that the resulting partition $\{ I_i \}$ 
has the property that
each $I_i$ is a subset of either $I$ or $J$.
By the homomorphism $h$ and its subalgebra property, there exists a tuple
$(t, x) \in R^{\relb}$ such that
$t \in T$ and $x_j \notin \maxscc{s(a_j)}$ for all $j \in J$ (just multiply all tuples in $R^\relb\cap (h(a_1) \times \cdots \times h(a_{k+l}))$).
By the fact that $T \subseteq \pr_I W$, we have a tuple
$(t, u) \in W$.  By the strong connectedness of $W$
(Proposition~\ref{prop:subdir-of-simple-sc}), 
there is a tuple $(t', u') \in W$ that is distinct from $(t, u)$
at each coordinate in $\cup I_i$ and such that $(t', u') \mul (t, u) = (t, u)$.
We also have $(t', u') \mul (t, x) = (t, u')$;
note that $u' \mul x = u'$ by conservativity of $\mul$.
  As $u$ and $u'$
differ at every coordinate in $J\cap(\cup I_i)$, the claim follows.

As a consequence of this last result, for any tuple
$t \in \pr_I W$ and any tuple $u \in \pr_J W$, it holds that
$(t, u) \in W$.
Further it holds that
$(\pr_I R^{\relb}\cap (h(a_1) \times \cdots \times h(a_k)))\cap (h'(a_1) \times \cdots \times h'(a_k))$ is a subdirect
product of $h'(a_1), \ldots, h'(a_k)$ (Lemma~\ref{lemma:subdir-scc}), and we have that
$h'$ is a homomorphism from $\rela$ to $\pow(\relb)$.
\end{pf}

\begin{pf}
(Theorem~\ref{thm:twosem-sac})
Suppose that $(\rela, \relb)$ has the SACC.
By Proposition~\ref{prop:sacc-implies-subalgebra-sacc},
the instance $(\rela, \relb)$ has the subalgebra SACC.
By Theorem~\ref{thm:sacc-on-max-components},
$(\rela, \relb)$ has the SACC via a mapping $s'$
where for all $a \in A$, it holds that
$s'(a)$ is a strongly connected subalgebra.
By assumption, each such $s'(a)$ is simple, and it follows
from Propositions~\ref{prop:subdir-of-simple-sc}
and~\ref{prop:sacc-almost-trivial}
that there is a homomorphism from $\rela$ to $\relb$.
\end{pf}

\section{Discussion}

In this work, we performed a systematic study of arc consistency and
three simple, natural extensions thereof.  We performed a comparison of
the studied consistency notions based on constraint languages, and proved
positive tractability results for singleton arc consistency.

Atserias and Weyer~\cite{atserias-weyer}
gave a uniform treatment of AC, PAC, SAC,
and general  consistency. 
Among other results, they show that it can be decided,
given a constraint language and
any pair of the previous consistency methods,
whether it is true that the set of instances that passes one of the
consistency tests coincides with the set of instances that passes the
other. Their results combined with the fact that general consistency/bounded width
is decidable~\cite{bkbw} implies that it can be decided whether or not a given
constraint language is solvable by any of the other methods.

We conclude by posing one question for future work.
Barto and Kozik~\cite{bkbw} have recently characterized all languages solvable
by bounded width.  Can all such languages be solvable by singleton arc consistency,
or are there bounded width languages not solvable by singleton arc consistency?
Resolving this question in the positive would seem to yield an interesting
alternative characterization of the bounded width languages.

\paragraph{Acknowledgements.}  
The authors thank Manuel Bodirsky for his comments and collaboration.
The authors also thank Johan Thapper for his many useful comments.

\bibliography{local}
\bibliographystyle{abbrv}

\end{document}